# Using Deep Learning and Machine Learning to Detect Epileptic Seizure with Electroencephalography (EEG) Data

**Haotian Liu**
Northeast Yucai Foreign Language School
Shenyang, Liaoning, China
Email: 2312938389@qq.com

**Lin Xi**
Northeast Yucai Foreign Language School
Shenyang, Liaoning, China
Email: 939791667@qq.com

**Ying Zhao***
Department of Engineering Science and Applied Math
Northwestern University
Evanston, IL, U.S.A
Eric940825@gmail.com

**Zhixiang Li***
Department of Biomedical Engineering
Shenyang Pharmaceutical University
Liaoning, China
106040205@syphu.edu.cn

**Abstract:**

The prediction of epileptic seizure has always been extremely challenging in medical domain. However, as the development of computer technology, the application of machine learning introduced new ideas for seizure forecasting. Applying machine learning model onto the predication of epileptic seizure could help us obtain a better result and there have been plenty of scientists who have been doing such works so that there are sufficient medical data provided for researchers to do training of machine learning models.

**Keywords:**

Epileptic Seizure Detection, Machine Learning, Deep Learning, Electroencephalography, Convolutional Neural Network, Recurrent Neural Network

# Contents

## 1. Introduction:

The prediction of epileptic seizure has always been extremely challenging in medical domain. However, as the development of computer technology, the application of machine learning introduced new ideas for seizure forecasting. Applying machine learning model onto the predication of epileptic seizure could help us obtain a better result and there have been plenty of scientists who have been doing such works so that there are sufficient medical data provided for researchers to do training of machine learning models. In our research, we applied traditional machine learning algorithms, such as Linear SVM, Logistic Regression, KNN (K Nearest Neighbors), and Neural Networks, like CNN (Convolutional Neural Networks), RNN (Recurrent Neural Networks), and LSTM (Long Short-Term Memory), for prediction. The emphasis of our research is to compare the AUC (Area Under the Curve) and accuracy of various models. The research result indicates that machine learning has made epileptic seizure prediction an achievable reality. Although still in testing stage, the method possesses grate reference value. This prediction may help patients to repress epileptic seizure before symptoms can be noticed. Therefore, it allows in time treatment for patients, reduces work pressure for medical workers, and brings more effective control for epileptic. In the future, through the accumulation of data and perfection of hardware, machine learning can be broadly applied in industrial fields.

## 2. Related Works:

The Electroencephalography (EEG) is a widely used technology for measuring and monitoring human brains' activity. It is a noninvasive electrophysiology monitoring method, which recorded the electrical activity of human brain using the electrode placed along the scalp. Because of the voltage fluctuations caused by ion currents in brain neuronsIt, EEG can measure the activity of cerebral cortex. It is a graphic display of the voltage difference between both sides of brain along time. It represents the ratio measurement of activity. As a result, EEG data shown as a continuous time sequence waveform of extremely tiny voltage signal, always have micro wave amplitude. Although waveform recording is at cerebral cortex, it was influenced by the activity from deep subcortical.

In the research by Kiral-Kornek et.al [1], they use deep learning methods to predict seizure. First they use iEEG data train deep learning classifier to distinguish signals before and after seizure. And then they did the standard testing. Second, as in their paper, "classifier performance was tested on held-out iEEG data from all patients and benchmarked against the performance of a random predictor". Third, they modified the predicting system to adapt each patient's features. Hence, the prediction system was adjusted and tuned so that sensitivity or time in warning could be selected as priority by the patient. The result is this system can provide on time and functional seizure prediction.

Also, in the paper by Antoniades, et.al [2], they decided to build deep learning model with automatic learning features. In particular, they considered to use CNN as deep learning method. It proved that these deep learning models can automatically learned IED data. The resulting model provided insights for various types of IEDs within the group, and was invariant to time differences between the IEDs.

Moreover, the work by Thodoroff et.al [3] points out that they applied methods with a considerable amount of data. They trained deep neural network with EEG data for predicting the seizure. At the same time, they collected spectral, temporal and spatial information for analysis seizure. Mostly, they focused on the cross-patient study of predicting the seizure. The main advantage of this deep learning model is that can generalize from patient to patient very well.

From these previous studies we can see applying machine learning methods on predicting seizure have many benefits; however we still have not found which method is the most effective way or time efficient way of seizure prediction. And that leads to our study during this research.

## 3. Methods:

### 3.1 Dataset:

The dataset for this study is generated from the Epileptic Seizure Recognition Data Set under UCI Machine Learning repository [4]. The dataset contains 500 patients' 4097 electroencephalograms (EEG) readings over 23.5 seconds. The 4097 data points are reshaped into 23 chunks with each chunk including 178 voltage signals corresponding to the brain activity within one second. Consequently, this multivariate time series dataset has 11500 subjects with each subject having a brain activity label regarding 178 features. 5 kinds of brain activities are seen in the dataset. We randomly select an instance under each category and visualize it in Fig. The detailed description of each activity can be found below. We also denote each class an abbreviation name:

**Seizure:** Seizure activity is recorded.
**Tumor Area:** Subject is diagnosed with tumor, and the EEG is collected from these epileptic brain areas.
**Health Area:** Subject is diagnosed with tumor however the EEG is collected from non-epileptic brain area.
**Eyes Closed:** Collected from the healthy subject when the eyes are keeping closed.

**Eyes Open:** Collected from the healthy subject when the eyes are keeping open.

### 3.2 Pre-processing:

The dataset was pre-processed in the next step in order to downstream to the classifiers. We first investigate the missing data. As the data source is well organized and the EEG is carefully recorded, no missing data is found. As can be seen from Fig. 1, the voltage distribution of an EEG reviews inconsistent range among different classes. The lowest point of Seizure is much lower compared with other classes. We standardized the voltage into the interval of [-1,1].

Our objective is in two-fold. In one fold, we would like to recognize Seizure from other activities. This is a binary classification task [5]. We easily unified the labels other than Seizure into a single label. In the other fold, we aimed at validating whether machine learning and deep learning can be used in EEG recordings multi-labels classification. Hence, we kept the original label for this part. We also found that the dataset is balanced, which means each class is one-fifth of the overall. No further imbalanced-against method is needed in the multi-label task. For the binary classification, some techniques were applied to adjust the weights of majority class.

The entire dataset was stratified split into 8:2, training: testing. We applied different validation methods in machine learning and deep learning, which will be introduced later in the following sections.

### 3.3 Machine Learning Classifiers:

We employed 6 machine learning classification methods to build the predictive model. Following classifiers were experimented: 2 linear classifiers: Supply Vector Machine with linear boundary (denoted as LinearSVM) and Logistic Regression (LR); 2 ensemble classifiers: Random Forest (RF) and Gradient Boost Decision Trees (GBDT); 1 probabilistic classifier: Naive Bayes (NB) and K Nearest Neighbors (KNN). 5-fold cross-validation was preformed to avoid over-fitting and tune parameters.

**K Nearest Neighbors:** KNN is a very intuitive model. The sample is classified based on the labels of its K nearest neighbors. Another advantage of the model is time efficiency. There is almost no training time due to its principle. Euclidean distance was used to determine the neighbors [6].

**Logistic Regression:** Logistic regression is a type of generalized linear model [7]. In logistic regression, the model is always trying to find a hyperplane to distinguish positive instance from negative. We applied L2-regularization to avoid over-fitting. One-versus-All scheme was used in the multi-labels classification. Weights of majority class were adjusted to compensate the imbalance.

**LinearSVM:** An SVM model is a representation of the examples as points in space, mapped so that the examples of the separate categories are divided by a clear gap that is as wide as possible [8]. L2-regularization, One-versus-All were also applied. As there is non-probabilistic, we converted the distance to decision boundary to form a probabilistic output.

**Random Forest:** A random forest consists of multiple decision trees, which bootstrap from subset randomly sampled entire dataset, to reduce the probability of over-fitting. We also limited the feature amounts under the square root of the complete set. Given that random forest classifier has many parameters to tune, we only tune two mainly factors: amount of estimators and max depth of each tree.

**Gradient Boosting Decision Tree:** Gradient Boosting is another method to ensemble decision trees. GBDT focus on using gradient optimization to reduce the error of previous generated parameters. Compared with random forest, GBDT is expected to have faster speed as it usually builds trees with smaller depth.

**3.4 Deep Learning Neural Networks:**

Recently, deep learning neural network such as convolutional neural network (CNN), recurrent neural network (RNN) and etc. show promising performance on one dimensional classification or prediction, such as speech recognition, stock price prediction. Therefore, we considered that using deep learning architecture to classify EEG recording. In this part, we will introduce 3 neural networks we deployed. As mentioned above, 20% has been set to be the held-out test set. We further stratified split training set into 8:2, to form a validation set. As our objective is to validate that deep learning can be useful in the EEG prediction, instead of pursuing the best model to yield a state-of-the-art result. We didn't make all of our efforts to design the configuration of layers, tune parameters.

**Convolutional Neural Network:** A Convolutional neural network (CNN) is a deep learning architecture that has multiple convolutional layers. CNN has been showing extremely promising usage in image classification, signal processing and etc. In the medical area, CNN presents its advantage in the support of decision making such as MRI image classification, drug relation detection. Other than the previous mentioned work, our input data is in 1-Dimension instead of 2 or 3-Dimensions. Although, recurrent neural network has a more common use in the 1-D classification task such as speech recognition, we explored whether a well-designed CNN is useful in seizure prediction with EEG. The architecture of our CNN is shown in Fig. 2. Input data, which is a 178-feature 1-D list, was first fed into the convolutional layer. Then, we used max pooling to select the most important features, followed by the second convolutional layers with more kernels. Another max-pooling layer was used after the second convolutional layer. Then 4 fully-connected layers were used to learn the non-linear combinations of previous layers [9]. 8 kernels were employed in the first convolutional layer, while we used 16 kernels in the second one [10]. Kernel size and batch size were tuned to yield a more competitive result.

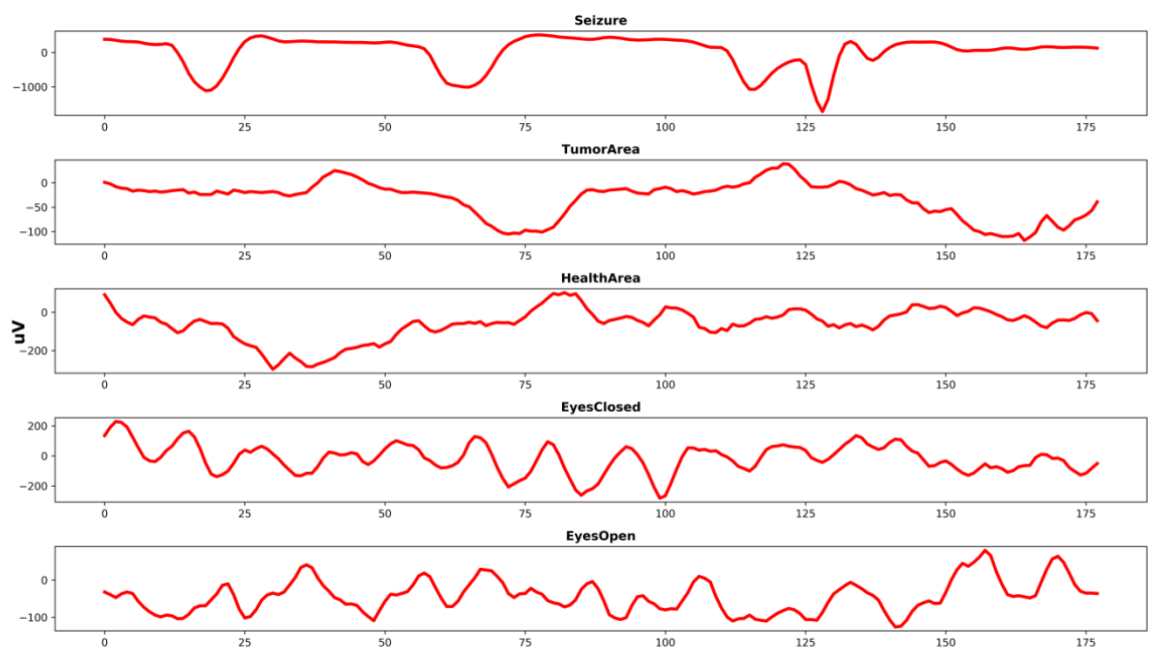

**Figure 1**: A demonstration of EEG recordings. An example of each brain activity was randomly selected and visualized. The activities from top to bottom are respectively: Seizure, Tumor Area, Health Area, Eyes Closed and Eyes Open.

**Recurrent Neural Network:** Recurrent Neural Network (RNN) is widely used in the classification or prediction with sequential data, such as stock price prediction and speech recognition. The strength of RNN is that it can 'memorize' the 'history' of sequential data and make accurate prediction. However, standard RNN network has drawback at vanishing gradient and incapable of memorizing long-distance input. Gated Recurrent Unit (GRU) and Long-Short Term Memory Network (LSTM) were designed to address the aforementioned drawback, respectively. In the experiment, we adopted both GRU and LSTM with same architecture. The only difference was how internal neurons were connected. We applied two layers of LSTM/GRU with 32 neurons followed by 2 fully connected layers to give classification results. A drop-out rate of 0.5 was set to avoid over-fitting. Batch size was also

32.

## 3.5 Evaluation:

In the binary classification, as the labels are imbalanced, besides accuracy, we adopted area under the receiver operating characteristic (AUC) to evaluate the performance. In the multi-label classification, we employed precision on each class and an overall averaged accuracy to estimate our classifiers.

## 3.6 Tools:

Our pipeline was built on Python version 3.6.3. Machine learning classifiers, cross-validation, data pre-processing, parameters tuning was implemented by Scikit Learn package. We designed the architecture of deep neural networks with PyTorch. All experiments were deployed on Google CoLaboratory, which has dual CPU kernels and a GPU to expedite the computation time.

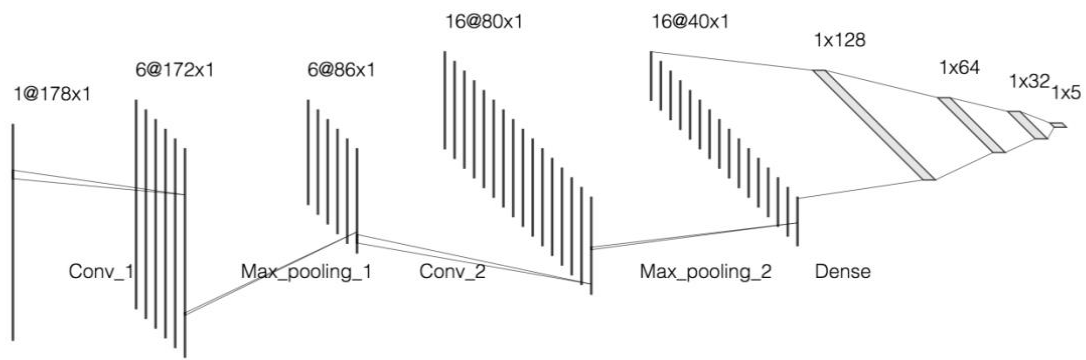

**Figure 2**: The CNN model. Architecture contained 11500-long 1-D list as input layers, 1-D convolution layers, fullyconnected layers, max-pooling layers and 5 probabilistic results as output layers.

**Table 1**: The performance of 6 machine learning classifiers and 3 deep learning networks on multi-label EEG classification. Precision on each label and average accuracy were reported. The best performance on seizure label and average accuracy were highlighted with bold-font.

|  | Seizure | TumorArea | HealthArea | EyesClosed | EyesOpen | Overall Accuracy |
|---|---|---|---|---|---|---|
| **Machine Learning** | | | | | | |
| Naive Bayes | 0.991 | 0.367 | 0.340 | 0.856 | 0.304 | 0.572 |
| Logistic Regression | 0.431 | 0.252 | 0.223 | 0.236 | 0.229 | 0.275 |
| LinearSVM | 0.347 | 0.242 | 0.188 | 0.257 | 0.211 | 0.251 |
| KNN | 0.991 | 0.367 | 0.340 | 0.856 | 0.304 | 0.572 |
| Random Forest | 0.935 | 0.652 | 0.571 | 0.632 | 0.355 | 0.629 |
| GBDT | 0.966 | 0.574 | 0.522 | 0.773 | 0.632 | 0.695 |
| **Deep Learning** | | | | | | |
| CNN | 0.901 | 0.574 | 0.642 | 0.707 | 0.663 | 0.703 |
| GRU | 0.962 | 0.688 | 0.450 | 0.779 | 0.641 | 0.704 |
| LSTM | 0.943 | 0.603 | 0.721 | 0.673 | 0.682 | **0.724** |

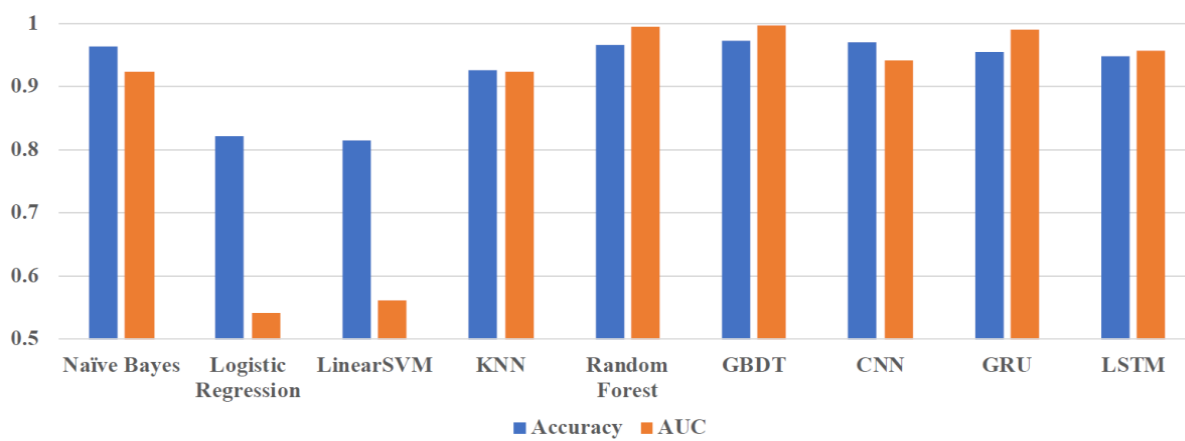

**Figure 3:** The performance of 6 machine learning classifiers and 3 deep learning networks on seizure recognition. Height of blue bars represents the accuracy, while yellow represent AUC. Y-axis had a cut-off as all values were over 0.5.

## 4. Results:

### 4.1 Results and Analysis:

In this section, we will report and compare the performance of 6 machine learning classifiers and 3 deep learning architectures on the held-out test set according to the evaluation methods mentioned in the previous section. As our experiments were 2-fold, Fig. 3 shows the results of binary classification, Tab. 1 demonstrates multi-label performance.

We first investigated results of binary classification. All 9 configurations yielded accuracy over 0.8. Except 2 linear methods, LR and linearSVM, all other had a competitive accuracy over 0.90. GBDT yielded the best performance of accuracy: 0.973 and AUC: 0.996, followed by RF's performance of accuracy: 0.965 and AUC 0.995. The ideal performance of these 2 classifiers inspired us that ensemble models of decision trees are a good fit for the seizure recognition. On the contrary, linear classifiers such as LR and linearSVM may not suitable for this task. As there were 11,500 features for one subject, linear models were easy to over-fit. The competitive results of both accuracy and AUC validated machine learning and deep learning models are useful in the seizure recognition.

We then revealed the performance of multi-label classification. Except, LR, linearSVM and KNN, all other configuration did a good work on distinguishing Seizure from other brain activities. They all yielded a high precision on Seizure of over 0.9. It is reasonable as all these configurations were well-performed in binary classification. Three deep neural networks all yielded an average accuracy over 0.70 while other machine learning classifiers didn't. GBDT had the best performance among machine learning classifiers with precision on Seizure: 0.966 and average accuracy 0.695. Three deep learning configurations had a higher average accuracy but were moderate at seizure recognition. The results demonstrated that if we were planning to recognize Seizure from all other brain activities, GBDT was the best fit. However, if we aimed at EEG classification, deep neural networks were the better candidate classifiers.

Next, we compared the difference among three deep neural networks. GRU and LSTM outperformed CNN on EEG classification in both seizure precision and average accuracy. It is reasonable as these two recurrent networks are good at handling time series data and could take both long and short memory data into consideration. However, these two recurrent networks were highly time-consuming. CNN spent less than 30 seconds to yielded this result. Although we deployed our configuration on Google Colaboratory with GPU on back-end, both recurrent networks need more than 30 minutes training time. We admitted that the results of our deep networks were biased due to the limitation of computational power. We didn't spend enough efforts in parameter tuning and architecture design. While, current results still demonstrated that deep neural networks were better approaches to EEG classification compared with tradition machine learning algorithms.

**4.2 Limitations and Future Work:**

There are also some limitations in our work that can be improved in the future. First, we aimed at recognizing epileptic seizure in a one-second time range. However, in real study, it has more clinical meaning to identify seizure activity in real-time base. Second, we didn't make abundant efforts in the refinement of neural network. If the deep learning architecture is better designed, a better performance in multi-label classification is expected. Moreover, engineering features are widely involved in the EEG recording interpretation. We are planning to extract more engineering features, such as characteristics in frequency domain. A combination of features in both time and frequency domain can convey more information towards the classifiers. Finally, we plan to enlarge the population of our dataset to yield a more robust, reasonable and reliable result.

**5. Conclusion:**

    Our work demonstrates that machine learning classifiers and deep neural networks are useful in the recognition of Epileptic Seizure from EEG recording. Specifically, 2 ensemble classifiers, random forest and gradient boosting decision trees yield the best performance of both AUC and accuracy over 0.95. Deep neural networks significantly outperformed machine learning classifiers in the multi-label classification of brain activities. Our models showed potential usage in clinical decision making, such as identifying seizure in a timely manner. Further study is needed to refine our networks to achieve a state-of-the-art result. A larger dataset is also important to validate the robustness of our models.